# Robots in the Garden: Artificial Intelligence and Adaptive Landscapes

Zihao Zhang[1], Susan L. Epstein[2], Casey Breen[3], Sophia Xia[4], Zhigang Zhu[5], Christian Volkmann[6]

[1]City College of New York/CUNY, New York/USA · zzhang@ccny.cuny.edu
[2]Hunter College/CUNY, New York/USA · susan.epstein@hunter.cuny.edu
[3]City College of New York/CUNY, New York/USA · cbreene000@citymail.cuny.edu
[4]Hunter College/CUNY, New York/USA · sophiax140503@gmail.com
[5]City College of New York/CUNY, New York/USA · zzhu@ccny.cuny.edu
[6]City College of New York/CUNY, New York/USA · cvolkmann@ccny.cuny.edu

**Abstract:** This paper introduces ELUA, the Ecological Laboratory for Urban Agriculture, a collaboration among landscape architects, architects and computer scientists who specialize in artificial intelligence, robotics and computer vision. ELUA has two gantry robots, one indoors and the other outside on the rooftop of a 6-story campus building. Each robot can seed, water, weed, and prune in its garden. To support responsive landscape research, ELUA also includes sensor arrays, an AI-powered camera, and an extensive network infrastructure. This project demonstrates a way to integrate artificial intelligence into an evolving urban ecosystem, and encourages landscape architects to develop an adaptive design framework where design becomes a long-term engagement with the environment.

**Keywords:** Resilience, climate change, technology, robotics, machine learning, artificial intelligence

## 1　Introduction

In the discipline of landscape architecture, a major epistemological framework has assumed that the environment is a closed and static system that can be measured, predicted, and conceivably controlled by technology (LYSTRA 2014). The reality of severe climate change challenges this view. Although advancements in industrial technology have given humans some control over their environment, carbon continues to be released into the atmosphere at unprecedented rates. While science rigorously measures and predicts the increasingly grim impact of human decisions, quantities of computer-processed data and complex control policies have yet to provide straightforward solutions to climate change. This paper argues for a research paradigm where artificial intelligence (AI) helps adapt landscapes to a changing environment rather than control them. It considers the role AI can play within this new focus on adaptivity, and how they can contribute to an adaptive design framework that requires a long-term engagement with the environment.

ELUA, the Ecological Laboratory for Urban Agriculture, offers a case study for an evolving ecosystem, embedded with AI, that responds to the uncertainties of a changing climate. In a collaborative endeavor among computer scientists, landscape architects and architects, two commercial gantry robots and an extensive infrastructure support cultivation in two polyculture gardens (Figure 1). Our work includes construction and customization of the robots, incorporation of sensor arrays, an AI camera, and network infrastructure, as well as the design and construction of the garden beds.

In the past two decades, many landscape programs have built laboratories with machinery and a "lab culture" as both research and education infrastructure. Examples include Alexander Robinson's work at the Landscape Morphologies Lab of the University of Southern California(ROBINSON & DAVIS 2018); Bradley Cantrell and Xun Liu's work with hydromorphology tables at the University of Virginia and Harvard University (LIU 2020); Matthew Seibert's work at Milton Land Lab;[1] and Ilmar Hurkxken and Christophe Girot's Robotic Landscape work at ETH Zurich (HURKXKENS 2020). Each of these develops landscape laboratories that integrate physical spaces with customized tools and machinery.

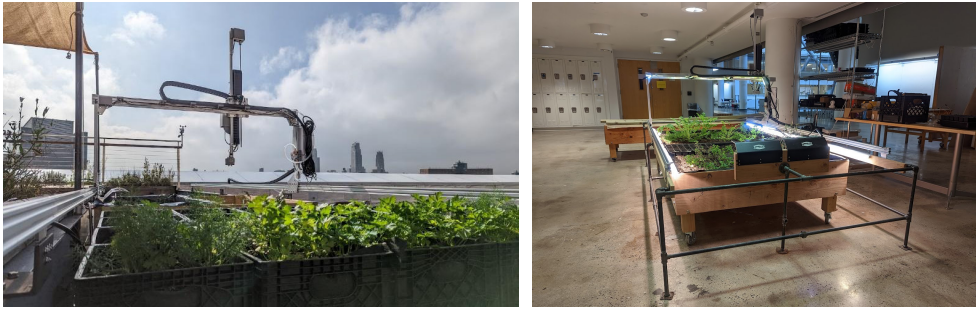

**Fig. 1:** Two robot-assisted gardens, outdoors (left) and indoors (right)

This phenomenon mirrors the 21st-century development in landscape theory that prioritizes dynamic landscape processes and ecological evolution over static forms. Landscapes are imagined to evolve with recursive and process-based strategies over time, instead of as a one-time construction. Projects such as the Fresh Kills and Downsview Park competitions in the early 2000s weres examples of this design paradigm (CZERNIAK 2001, REED & LISTER 2014). Since then, many scholars have incorporated a broad range of ideas and concepts from both sciences and humanities to diversify and develop that paradigm. They include multispecies co-production, novel ecology, feral ecology, and cyborg landscapes (HOUSTON, HILLIER, MACCALLUM, STEELE & BYRNE 2018, KLOSTERWILL 2019, PROMINSKI 2014). In addition, new tools have been imagined for integration into landscape systems that would execute process-based strategies and co-evolve with other landscape actors (CANTRELL & HOLTZMAN 2015).

Our research contributes to this body of work in theory and practice. We view an intelligent system, like its human counterparts, as an imperfect agent, rather than an omniscient, omnipotent black box. The perspective of collaborative intelligence (EPSTEIN 2015) provides an emergent, constructive view of artificially intelligent agents that participate in and support a collaborative design process. We envision an alternative future where technology plays a more integral role in adaptation to rapidly changing environments.

---

[1] https://miltonlandlab.cargo.site/About

This paper documents the design, construction, and preliminary testing of ELUA, and provides practical recommendations for such landscape laboratories. It also reflects on the ramifications of ELUA for landscape design and argues for a new research paradigm where AI is an integral part of evolving ecosystems. From our perspective, landscape design is no longer a finished product, but a long-term engagement and collaboration with an assemblage of actors, including AI systems, that co-creates an evolving ecosystem.

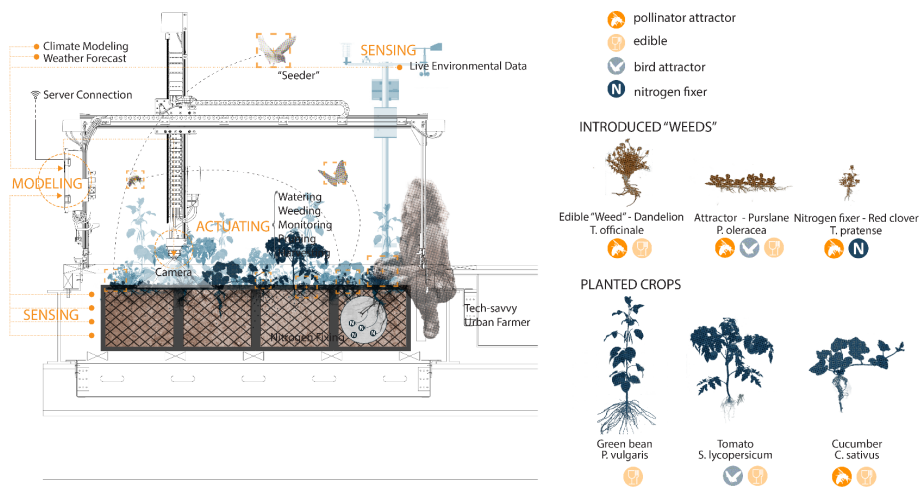

**Fig. 2:** Diagram of a robot-assisted polyculture garden where AI agents are considered part of an evolving ecosystem (© 2022 Z. Zhang, C. Breen and C. Volkmann)

## 2   Technologies and Design-build

ELUA has two sites within the Spitzer School of Architecture at City College of New York, an outdoor garden on a rooftop and an indoor garden in a communal area near the landscape studios. The outdoor garden (shown in Figures 1 and 2) focuses on growing food; the indoor garden (shown in Figures 1 and 3) is used for education, prototype research, and experimental development.

### 2.1   Initial Hardware and Software

Both ELUA robots are from FarmBot,[2] a California-based firm that designs and markets open-source commercial gardening robots, and develops web applications for users to interface with those robots. These are gantry robots that operate in three dimensions and employ interchangeable tool heads to rake soil, plant seeds, water plants, and weed. FarmBots are highly customizable; users can design and replace most parts to suit their individual needs. For ELUA, we have designed and 3D-printed our own watering nozzles, seeders, seed troughs, and camera mounts.

---

[2]  https://farm.bot/

FarmBot's supporting code for farm design and robot control is also open source; users can customize it through an online web app. This allows us to revise or replace the provided executable programs and to introduce new functionality into ELUA. The basic FarmBot code visualizes garden designs before planting, photographs the garden, and provides primitive sensing and behaviors.

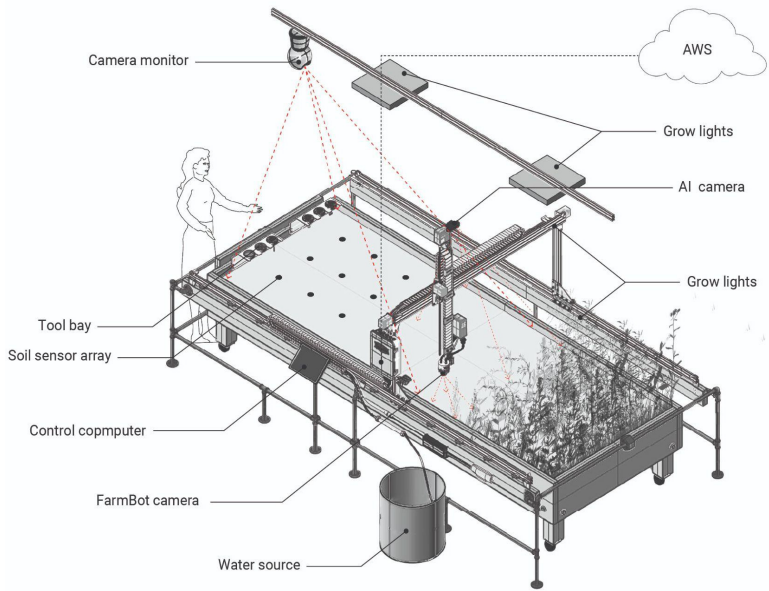

**Fig. 3:**   The indoor gardening system (© 2022 Z. Zhang and C. Breen)

## 2.2    Customization for Robot-assisted Gardening

We customized each of ELUA's robots in several ways for our indoor and outdoor gardens, and both systems function as intended. ELUA's rooftop robot has 2-meter tracks from a third-party vendor tailored to the spatial parameters of its site; they replace FarmBot's original (*x*-axis) 1.5-meter robot tracks. To install the tracks on the I-beams on the rooftop, we designed and built our own joints. We developed planters made with standard milk crates lined with a layer of geo-fabric. This modular approach provides flexibility to the entire rooftop design and installation. The 5'x5' structural frames are custom-built with 10-foot, 16-gauge steel drywall studs and tracks, cut to size and assembled on-site with L-shaped corner clips. The structural frames fit between two I-beams and support milk crate planters or wooden planting boxes (Figures 1, 2 and 4). As shown in Figures 1 and 3, the indoor system consists of a black-pipe armature for the robot and mobile garden beds, instead of gantry tracks fixed directly onto the garden beds as suggested by FarmBot Inc.. This armature design allows us to remove and replace mobile garden beds if needed without deconstructing the entire gantry.

The rooftop garden will eventually be used by a student group to produce food. In contrast, the indoor garden is intended for more advanced experiments, where we will prototype and develop new algorithms, tool heads, and operations to be used for both robots.

## 2.3 Computer Vision and Garden Maps

Each FarmBot includes a camera that photographs the garden, with software to roughly stitch the images together. We developed algorithms to improve image mosaicing (DICKSON et al. 2002, MOLINA & ZHU 2014). Meanwhile, we installed a second, more powerful AI camera OAK-D camera[3] to perform more advanced computer vision tasks, such as depth detection, weed detection and plant identification. We have used this AI camera and developed seamless image stitching for two-dimensional aerial views in ELUA that are more accurate and more visually appealing.

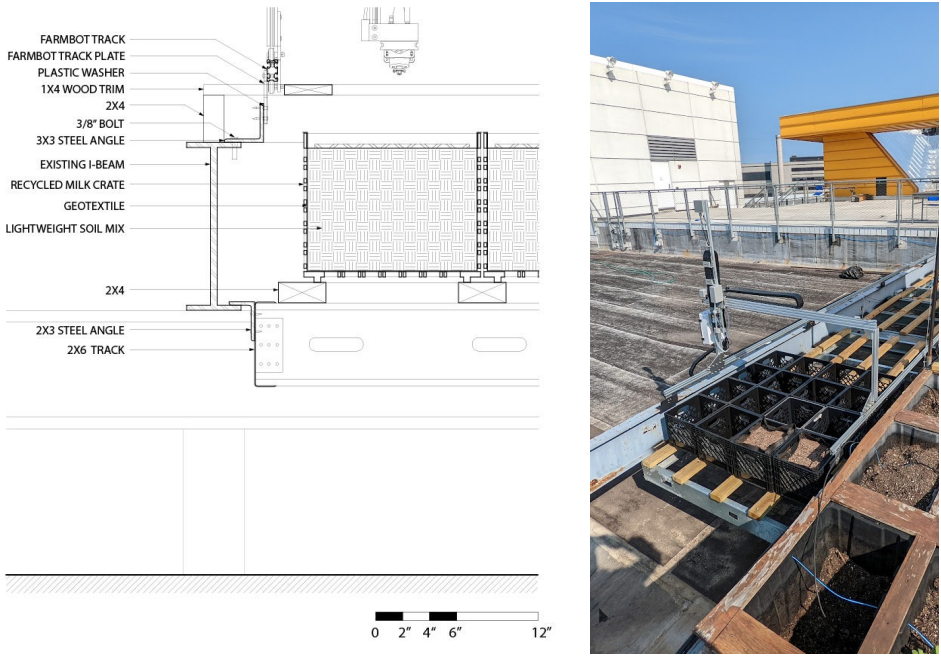

**Fig. 4:** Construction details for the rooftop robot garden (left, © 2022 Z. Zhang and C. Volkmann) and a photo during construction of the rooftop garden beds (right)

Initially, the user describes the garden's contents to a FarmBot as a simple placement of plants from the provided "plant dictionary" on a *garden map,* a two-dimensional grid visualized by the web app. FarmBot stores the location of each plant as a datapoint (*x, y*) on that map. Other emerging plants, if detected by the camera, are treated uniformly as "weeds" that should be managed by the robot. FarmBot's software has no plant identification algorithm to differentiate between different weed species. Some "weeds," however, such as dandelion and purslane, are edible, while others, such as red clover, can fix nitrogen and support soil health. We expect that these species could play important roles in an urban polyculture garden and increase urban biodiversity and resilience. Thus, we intend to process images from the AI camera with deep learning models to detect such opportunistic species, record their locations in the garden map, and have the robot cultivate all welcome but unanticipated plants.

---

3   OpenCV AI Kit: https://store.opencv.ai/products/oak-d

## 2.4   Multimodal Sensing and the New Database

In addition to the FarmBot armature, our gardens are designed to benefit from additional sensors. We have incorporated an array of capacitive soil-moisture sensors connected to a microcontroller with a WiFi module. Our outdoor garden also includes a personal weather station connected to Weather Underground. With their application programming interface (API) service, we can access real-time and historical weather data, as well as a seven-day weather forecast from the rooftop. Additional sensors could be similarly installed to measure other environmental factors, such as solar radiation, $CO_2$, and air pollution.

To incorporate this sensor data into ELUA, we have created a virtual server that hosts our own database as well as any API services. This greatly expands ELUA's capability because it connects each FarmBot to other types of open data and services. For example, with weather forecast data, ELUA could modify scheduled watering regimens for precipitation and drought.

## 2.5   Machine Learning and AI

AI is pervasive in this research. Non-experts. including many landscape researchers, often think of an AI system as a general artificial intelligence that addresses multiple goals simultaneously. In ELUA, however, AI algorithms are individually built for specific tasks.

In ELUA, the AI camera we added processes images with OpenCV, an open-source computer vision and machine learning software library. This provides machine learning algorithms, including pre-trained deep neural network modules that can be modified and used for specific tasks, such as measuring plant canopy coverage and plant height. Machine learning and AI planning can also be used with the multimodal sensory data described in Section 2.4 to provide data-driven guidance to improve garden management.

An AI system that relies on reinforcement learning (RL) develops a policy for its behavior when it is rewarded or punished for the outcome of its actions. Such systems have devised unexpected behaviors in Go, chess, and some video game that expanded human players' understanding of these games and provided new insights (SCHRITTWIESER et al. 2020). Landscape architects and ecologists now also imagine how RL systems might manage the environment and construct wild landscapes (CANTRELL et al. 2017, ZHANG & CANTRELL 2021). One research team conducted RL experiments to prune a polyculture garden with a FarmBot to increase biodiversity (PRESTEN et al. 2022). We will perform RL experiments in a simulated environment with a virtual robot and later test the learned policies with a physical robot. We envision a version of ELUA that will evolve and propose novel methods, such as combinations of different watering nozzles and watering paths in different scenarios. We hope some unexpected combinations will surprise, intrigue, and inspire us with their successful outcomes that help the garden adapt to a changing climate.

## 3   Results and Discussion

In the fall of 2022, we planted herbs on the rooftop and lettuce and herbs in the indoor garden to learn and test the basic functions of the robots. (The indoor ELUA robot in action appears in a brief video at https://youtu.be/VTec_SXO5Lk.) Both ELUA robots captured garden maps and carried out watering events as expected, although maintenance and troubleshooting are needed from time to time. This is a design-build project and every aspect of ELUA was constructed by faculty and students. Our team has gained hands-on knowledge as it constructed ELUA. Here, we offer three recommendations.

First, in cities like New York, public and free resources are available for academic research and well worth the time to track down the networks of organizations and groups. We received 3 cubic meters of free, clean mineral soil from the NYC Clean Soil Bank hosted by the NYC Mayor's Office of Environmental Remediation. The soil was delivered by the crews from the New York Restoration Project. We received 40 bags (1 cubic meter) of compost from the NYC Composting Project hosted by Big Reuse. In return, we took the Big Reuse crews to tour ELUA, and hope to maintain this relationship. We learned about our current soil mix (⅓ mineral soil, ⅓ compost, and ⅓ perlite) during a free guided tour of a green-roof facility hosted by the NYC Department of Parks & Recreation. ELUA would have been far more costly and difficult to construct without these public resources. We encourage prospective developers to seek out similar assistance.

Second, "open-source" offers adaptivity to its users but also forewarns the necessity of substantial troubleshooting. Although FarmBots appears to be user-friendly, some knowledge of computer science and electrical engineering is required to set up such a system. Researchers and research assistants from our Departments of Computer Science successfully overcame many issues during our installation. Moreover, a licensed architect on the team successfully designed and constructed the garden beds and installed a FarmBot onto the existing load-bearing rooftop structure. A multidisciplinary team of computer scientists and designers is highly recommended to replicate ELUA.

Finally, institutional knowledge is important in academic research. Institutional structures in universities, especially public schools, can support or hinder academic research visions. Researchers need to be nimble and adaptive in pursuing their goals. For example, thanks to the Architecture School, the Colleges, and the University, we received multiple seed funds for ELUA. The Architecture School also provided space and infrastructure to house ELUA. Some rules, however, could not be bent. Because the University's security regulations blocked network ports used by a FarmBot, we have had to set up alternative 5G Wi-Fi hotspots untill the University can provide a research-only network. We recommend an ample buffer in the research schedule to account for unexpected circumstances, as well as a good rapport with university offices.

## 4    Conclusion

Despite ELUA's practical focus on urban food production, it is also a thought experiment that challenges landscape architects' conventional views on agency and intelligence. With ELUA, we want to formulate a theory that questions how the environment is conceived and constructed. To some extent, ELUA offers a glimpse into an ecosystem of computerized ecology where the relationship between humans and plants is deeply mediated and, at the same time, enabled by sensors, controllers, computer hardware, layers of computer code, and online servers.

A problem in the perspective of landscape architecture is its current conception of AI as intelligence embodied in a single agent (CANTRELL & ZHANG 2018). Ideas in posthumanism, however, such as assemblage and sympoiesis become new concepts to reframe agency and intelligence as distributive (BENNETT 2010, HARAWAY 2016, TSING 2015). From this posthumanist perspective, intelligence should be viewed as an emergent epiphenomenon arising from the interactions of an assemblage of actors – humans, AI agents, animals, and plants. This framing sheds light on a new landscape design paradigm based on co-evolution among biotic and abiotic agents. We could allow AI systems and plant and animal agents to co-evolve to create novel ecosystems that inspire us with new methods to construct the landscape.

In this new landscape design paradigm, AI agents would no longer simply model human behavior under human control. Instead, they would become co-creators among human and nonhuman actors. They could offer novel approaches and long-term cultivation strategies that humans can learn from and use to adapt to the changing climate. ELUA is a physical demonstration of this new paradigm of landscape design. It provides empirical evidence that collaboration among AI agents and other human and nonhuman actors is within reach.

# Acknowledgment

This research is partly funded by a CUNY Interdisciplinary Research Grant. The authors are ranked in descending order of contribution to the paper. All authors contributed to the conceptualization of the project. Z. Zhang. oversaw the project, assisted by C.B. S.L.E. led robot behavior, database, sensing, machine learning, and simulation development. S.X. developed foundational material for communication, scheduling, and computer vision; Z. Zhu led preliminary computer vision, machine learning, and simulation development. C.V. led the design-build of the rooftop garden. The authors thank the many research assistants who have contributed to this project. Kevin Lin provided hardware and systems expertise, and Ghazanfar Shahbaz constructed the database. Nicole Girdo, Kaitlin Labatt, and Aparna Ramesh, and Gaël Oriol constructed and installed the rooftop garden. Donald Lushi and Albi Arapi conducted computer vision experiments with the FarmBot camera and installed and tested the AI camera. Gong Qi Chen conducted the RL experiment.